%% file: main.tex
\documentclass[letterpaper]{article}
\usepackage{grounded_dreamer_preprint,times}

\usepackage{iftex}
\ifPDFTeX
  \pdfoutput=1
  \usepackage[utf8]{inputenc}
\else
  \AtBeginDocument{}
\fi
\usepackage[T1]{fontenc}
\usepackage{hyperref}
\usepackage{url}
\usepackage{booktabs}
\usepackage{amsfonts}
\usepackage{nicefrac}
\usepackage{microtype}
\usepackage{xcolor}
\usepackage{graphicx}
\usepackage{amsmath}
\usepackage{tabularx}

\let\cite\citep

\title{Grounded-Dreamer: Robot Learning from World Model Synthetic Data with Physical Grounding}

\author{
    \textbf{Feng Jiang}\textsuperscript{1*},
    \textbf{Shu Chen}\textsuperscript{2*},
    \textbf{Yuanfei Wang}\textsuperscript{1*},
    \textbf{Yuan Li}\textsuperscript{1}, \\[0.3em]
    \textbf{Zhenhao Shen}\textsuperscript{1},
    \textbf{Jiaqi Liang}\textsuperscript{1},
    \textbf{Yang Chen}\textsuperscript{1},
    \textbf{Lei Wei}\textsuperscript{1},
    \textbf{Chen Xie}\textsuperscript{3},
    \textbf{Ruihai Wu}\textsuperscript{1\ensuremath{\dagger}} \\[0.6em]
    \textsuperscript{1}Peking University \\
    \textsuperscript{2}Hong Kong University of Science and Technology (Guangzhou)\\
    \textsuperscript{3}Lightwheel\\[0.5em]
    \textsuperscript{*}Equal contribution \quad
    \textsuperscript{\ensuremath{\dagger}}Corresponding author \\
    Contact email: \href{mailto:wuruihai@pku.edu.cn}{\texttt{wuruihai@pku.edu.cn}}
}
\hypersetup{
    pdftitle={Grounded-Dreamer: Robot Learning from World Model Synthetic Data with Physical Grounding},
    pdfauthor={Feng Jiang, Shu Chen, Yuanfei Wang, Yuan Li, Zhenhao Shen, Jiaqi Liang, Yang Chen, Lei Wei, Chen Xie, Ruihai Wu},
    pdfsubject={Robot learning from physically grounded synthetic data},
    hidelinks
}

\begin{document}
\maketitle

\input{sections/0_abs}
\input{sections/1_intro}

\input{sections/2_related}

\input{sections/3_method}
\input{sections/4_exps}

\bibliographystyle{iclr2027_conference}
\bibliography{main}

\clearpage
\appendix
\input{supp}

\end{document}

%% file: sections/0_abs.tex
\begin{abstract}
Scaling robot manipulation policies requires large amounts of diverse, action-labeled data, yet collecting such data in the real world remains expensive and difficult to scale.
Existing methods either depend on human teleoperation, rely on simulation with limited visual realism, or use human videos that lack embodiment-specific annotations. 
In this paper, we introduce Grounded-Dreamer, a framework for synthesizing physically grounded robot manipulation data with video world models while avoiding human teleoperation demonstrations. 
Grounded-Dreamer combines the visual realism of generative video models with simulator physics engines. 
Starting from a small set of real images depicting robot task initial states, our method automatically reconstructs task-relevant objects and camera geometry, aligns them in simulation, and collects physically valid trajectories. 
These trajectories are used to post-train both a video world model and an inverse dynamics model, enabling the generation of visually realistic robot manipulation videos and corresponding robot actions from real images. 
To filter physically implausible rollouts, Grounded-Dreamer replays generated trajectories in the reconstructed simulation and retains only those that successfully complete the tasks. 
The real-world robot policies are trained solely on the grounded synthetic trajectories without any human demonstrations.
Empirical results in real-world tasks demonstrate that Grounded-Dreamer can synthesize physically grounded data and substantially facilitate policy learning without requiring human teleoperation effort.

{\footnotesize\noindent Project Page: \url{https://fffstrong.github.io/Grounded-Dreamer/}.\par}
\end{abstract}

%% file: sections/1_intro.tex
\begin{figure}[htbp]
  \centering
  \includegraphics[width=\linewidth]{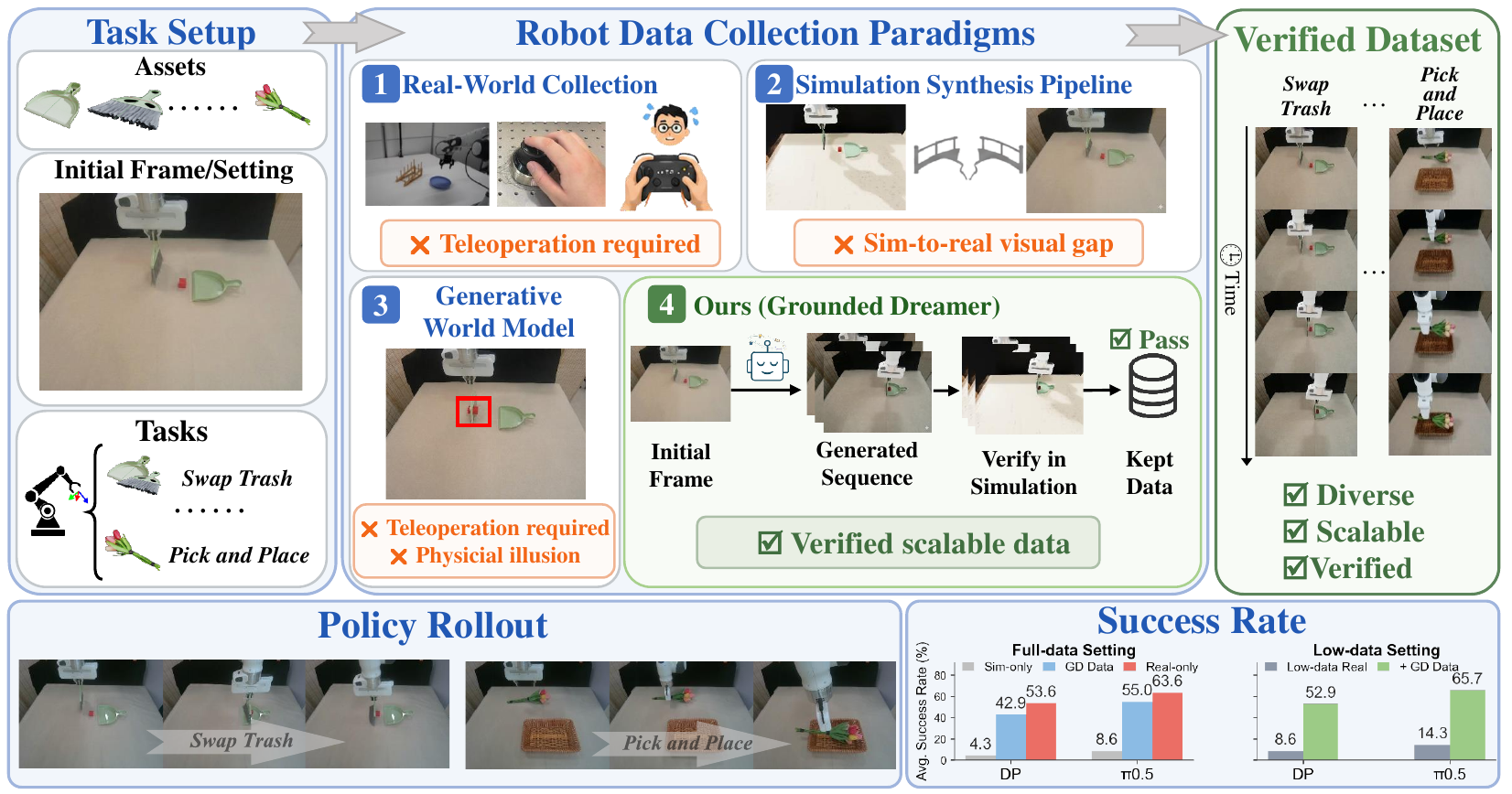}
  \caption{\textbf{Overview of Grounded-Dreamer.} 
  Our framework automates the generation of massive, high-fidelity robot data from  a small number of real-world reference images. By combining generative world models with simulation-based verification, Grounded-Dreamer produces data that is both visually realistic and physically consistent. This simulation verification loop effectively filters out the physical hallucinations common in purely generative approaches, allowing robot policies to be trained directly on scalable synthetic data that transfers robustly to complex real-world scenarios.    
  }
  \label{fig:teaser}
  
\end{figure}

\section{Introduction}
\label{Intro}

Building general-purpose robot policies capable of performing diverse manipulation tasks across varied environments requires large-scale, high-quality data. However, robot data is difficult to scale, and existing sources present a challenging trade-off between quality and quantity. Real-world teleoperation provides high-fidelity demonstrations but is limited by intensive human labor. While UMI data collection~\cite{chi2024universal, xu2025dexumi} improves efficiency, there remains a need for manual operations. Conversely, simulation data offers inexpensive, action-labeled trajectories but is hindered by the sim-to-real visual gap. Human-centric videos are abundant and diverse, yet they lack the robot action annotations essential for direct policy training. Consequently, the field remains trapped between costly manual collection and the limitations of imperfect proxy data.

This dilemma motivates the pursuit of scalable robot data synthesis. Prior efforts such as DemoGen~\cite{xue2025demogen} and MimicGen~\cite{mandlekar2023mimicgen} have attempted to augment existing datasets by diversifying robot demonstrations; however, these methods still rely on human-collected trajectories as initial seeds. Other approaches, including RoboGen~\cite{wang2023robogen} and GenSim~\cite{wang2023gensim,hua2024gensim2}, leverage Large Language Models (LLMs) and simulators to automate task generation and trajectory collection. While these frameworks significantly reduce human intervention, the resulting data remains confined to simulation, leaving the visual sim-to-real discrepancy unresolved.

Generative video world models offer a potential alternative. Trained on internet-scale data, these models capture rich visual and world dynamics priors, enabling the generation of manipulation rollouts. When paired with an Inverse Dynamics Model (IDM), such videos can be converted into action-labeled trajectories. However, pretrained models like Cosmos~\cite{agarwal2025cosmos} are primarily trained on human-centric videos. Directly applying them to robotic tasks often results in "hallucinated" physics, such as inconsistent object motions or non-physical contacts.

In this work, we introduce \textbf{Grounded-Dreamer}, a framework that synthesizes high-fidelity robot manipulation data without a single human teleoperation demonstration. Our core insight is to link the visual realism of generative world models with the grounding of physics simulators. We first build an automatic real-to-sim pipeline that calibrates camera poses, reconstructs task-relevant objects, and aligns them with the scene. Within this reconstructed simulator, we automatically collect diverse trajectories by leveraging the simulator's ground-truth states. These grounded trajectories serve as the foundation for post-training both the video world model and the IDM.

During data synthesis, the post-trained world model generates visually realistic manipulation videos from real-world initial images, while the IDM extracts the corresponding robot actions. To guarantee physical validity, we implement a closed-loop verification step: replaying the inferred trajectories within the reconstructed simulation. This process filters out physically inconsistent generations, ensuring that only successful, grounded trajectories are used for further policy training.

We evaluate Grounded-Dreamer across seven diverse real-world manipulation tasks, spanning short-to-long horizon sequences and sophisticated tool use. Our empirical results demonstrate that policies trained exclusively on grounded synthetic data achieve performance competitive with those trained on real-world human demonstrations. Furthermore, owing to the inherent visual diversity and our targeted augmentation strategies, Grounded-Dreamer significantly outperforms real-data policies in out-of-distribution scenarios, showcasing robust zero-shot generalization across novel scenarios.

Our contributions are summarized as follows:
\begin{itemize}
\item We propose \textbf{Grounded-Dreamer}, a data synthesis framework that eliminates the need for human demonstrations by linking video world models with simulator physics engines.
\item We introduce an automated real-to-sim reconstruction and collection pipeline, enabling data synthesis that maintains both high visual fidelity and strict physical grounding.
\item We demonstrate through extensive real-world experiments that our grounded synthetic data effectively facilitates policy learning, achieving robust performance across diverse scenarios through purely autonomous data generation.
\end{itemize}

%% file: sections/2_related.tex
\section{Related Work}

\subsection{Robot Data Synthesis}
 
Simulation has been widely used to generate diverse conditions for robot data. Early sim-to-real work relied on visual or dynamics randomization \citep{tobin2017domain,peng2018sim}. Later environments extended this idea to broader robot-learning settings: RLBench standardizes manipulation tasks \citep{james2020rlbench}, robosuite and SAPIEN provide modular robot-object simulation platforms \citep{zhu2020robosuite,xiang2020sapien}, and Isaac Gym and RoboCasa emphasize scalable physics or household manipulation \citep{makoviychuk2021isaacgym,nasiriany2024robocasa}.

Recent generative frameworks, such as
GenSim, GenSim2, and RoboGen, use language or foundation models to generate tasks, scenes, demonstrations, or supervision \citep{wang2023gensim,hua2024gensim2,wang2023robogen}. Their data quality, however, still depends on generated task programs, planners, solvers, or simulator assumptions. For manipulation, local grounding outweighs global visual realism. Incorrect object pose, axis, or camera frame can make a plausible trajectory physically misleading. 6D pose estimation and reconstruction recover task-relevant structure from real observations \citep{hodan2018bop,wen2024foundationpose}, while structure-from-motion, NeRF, and 3D Gaussian Splatting facilitate view-consistent geometry or appearance \citep{schoenberger2016sfm,mildenhall2020nerf,kerbl2023gaussian}.

Another route reuses collected trajectories. A series of studies adapt human demonstrations to new object configurations \citep{mandlekar2023mimicgen}, corrective interventions~\citep{hoque2024intervengen}, long-horizon skills~\citep{garrett2024skillmimicgen}, and bimanual dexterous manipulation \citep{jiang2024dexmimicgen}. DemoGen further adapts one demonstration and edits the 3D scene to synthesize new observations \citep{xue2025demogen}. DROID and Open X-Embodiment instead scale real robot data directly, but require substantial collection effort \citep{khazatsky2024droid,oneill2024openx}. Grounded-Dreamer lies between these directions: it grounds synthetic data in real scene-object inputs, preserves physical structure through simulation, and expands the distribution with world-model rollouts.

\subsection{World Models for Robot Learning}

World models were originally developed as learned dynamics models for predicting future states before acting \citep{ha2018worldmodels,hafner2019dreamer,hafner2023dreamerv3}. In robotics, visual foresight leveraged video prediction for model-based control by selecting actions whose predicted futures match a goal \citep{finn2017visualforesight}. More recently, diffusion-based video models have enabled realistic rollout generation from images or text \citep{ho2022videodiffusion}, while systems such as UniSim and Genie move toward controllable visual worlds \citep{yang2023unisim,bruce2024genie}.

However, visually plausible videos are not directly usable robot training trajectories, as they often lack actions, embodiment constraints, and physically reliable interactions. DreamGen adapts video world models to robot learning by recovering pseudo actions through latent action or inverse dynamics models \citep{jang2025dreamgen}, while GigaBrain-0 further scales world-model-generated data for VLA training \citep{gigabrain2025}. Prior works have also explored inverse dynamics models to bridge observations and actions \citep{agrawal2016learning,pathak2018zeroshot}. Once converted into action-conditioned trajectories, such generated data can train policies including Diffusion Policy, RT-2, and $\pi_0$ \citep{chi2023diffusion,zitkovich2023rt2,black2024pi0}. Grounded-Dreamer builds on this line of work by coupling world-model rollouts with reconstructed scene inputs and simulation-based grounding to ensure both visual realism and physical consistency before policy learning.

%% file: sections/3_method.tex
\begin{figure}[htbp]
  \centering
  \includegraphics[width=\linewidth]{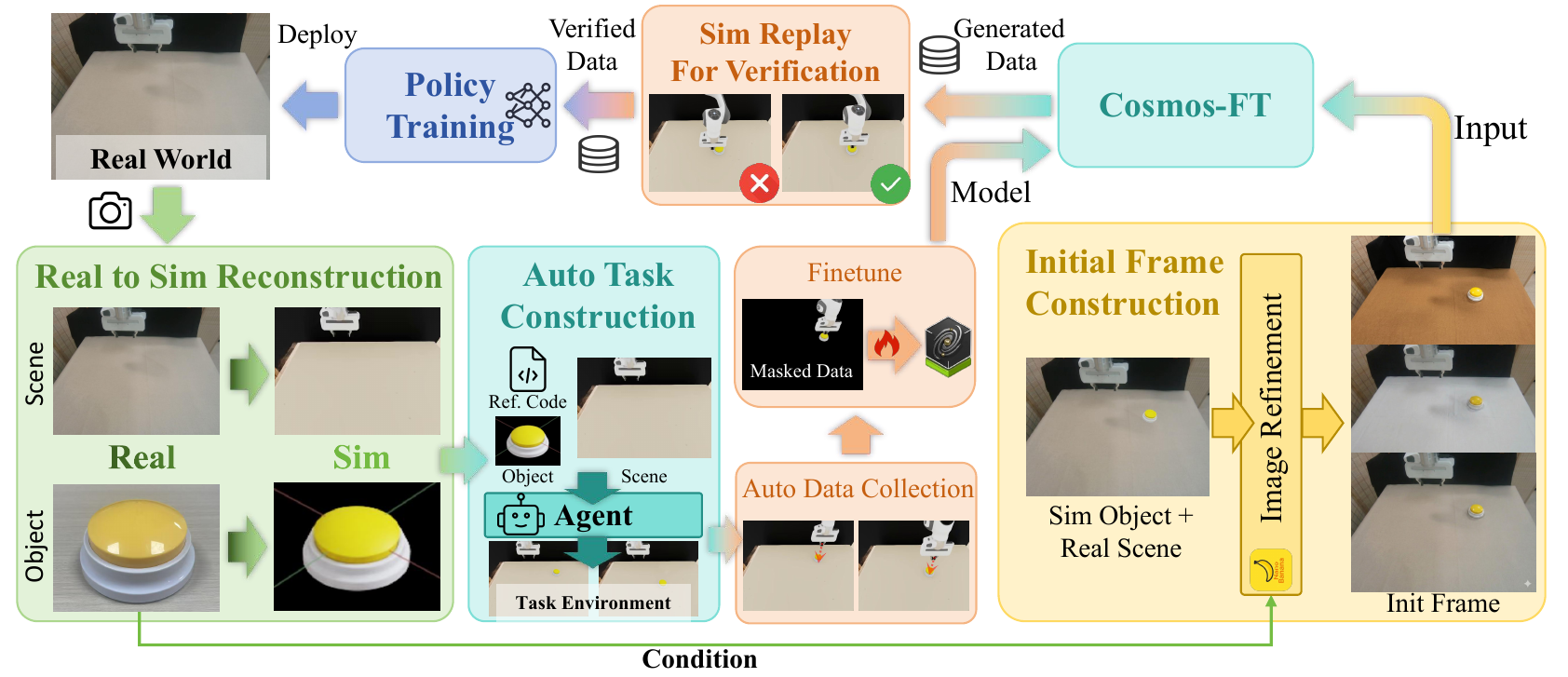}
      \caption{\textbf{The Grounded-Dreamer Pipeline.} Given several real-world reference images, our pipeline first reconstructs physically aligned simulation scenes and objects. 
      Within these environments, task data is automatically collected and processed with interaction-aware masking to fine-tune Cosmos-Predict 2.5 as the core data generation model. 
      An image refinement model then synthesizes diverse initial frames by combining reconstructed assets with real scene backgrounds and object appearances. Conditioned on these frames, the generation model produces manipulation videos, which are converted into action trajectories using an Inverse Dynamics Model (IDM). Finally, all synthesized trajectories are replayed in simulation for closed-loop verification, filtering out physically inconsistent samples before policy training and real-world deployment.
      }
  \label{fig:pipeline}
  
\end{figure}

\section{Methodology}
\label{sec:method}

We introduce \textbf{Grounded-Dreamer}, an autonomous pipeline designed to generate verifiable, high-fidelity synthetic data for scalable robot policy learning without the need for human-collected real-world trajectories. As illustrated in Figure~\ref{fig:pipeline}, starting from a small set of reference images captured in the real world, Grounded-Dreamer synthesizes massive, validated datasets through an automated workflow, which consists of four core stages: 
(i) \textbf{Task-Oriented Simulation Reconstruction}, where we reconstruct physically consistent simulation scenes and task-relevant objects; 
(ii) \textbf{Sim Data Collection and Interaction-Aware Masking}, where simulation data is automatically collected to fine-tune Cosmos-Predict 2.5 and the IDM with interaction-aware masking;
(iii) \textbf{Initial Frame Synthesis}, where the Nano Banana-based pipeline aligns real-sim composites and generates augmented initial images to enhance diversity and visual fidelity;
(iv) \textbf{Data Synthesis and Closed-Loop Verification}, where synthetic trajectories are generated from real-world initial frames and validated via simulation verification.

\subsection{Task-Oriented Simulation Reconstruction}
\label{subsec:reconstruction}

\paragraph{Scene Reconstruction.}
Our framework is built upon the Lehome simulation environment~\cite{li2025lehome}. Given the workspace boundaries and the relative pose of the robot base, we construct a lightweight simulation scene containing the robotic arm and workspace table. To align the simulated and real-world viewpoints, we perform extrinsic calibration using the FEEPE framework~\cite{wu2025foundation} and directly adopt the intrinsic parameters of the physical camera. This process ensures consistent fields of view (FOVs) between simulation and reality, providing a reliable geometric foundation for data grounding. For efficiency, we reuse off-the-shelf Lehome assets for tabletop textures and environmental backgrounds, since pixel-level fidelity in task-irrelevant regions is unnecessary for our pipeline.

\paragraph{Task-Oriented Object Reconstruction.}
For \textbf{rigid objects}, we employ SAM3D to reconstruct 3D geometry from the initial observation frame and rescale the resulting meshes to match real-world dimensions~\cite{chen2025sam}. For \textbf{articulated objects}, we first use SAM3D to decompose the object into individual components~\cite{chen2025sam}. Our autonomous agent then takes these components, together with predefined connectivity rules and joint specifications, and automatically assembles the articulated kinematic structure within the simulator.

\subsection{Sim Data Collection and Interaction-Aware Masking}
\label{subsec:masking}

With the task-aligned simulation environment established, we develop an automated data collection pipeline for trajectory generation necessary for world model post-training. Instead of manually scripting each task, we employ an LLM-based agent to synthesize and execute data collection scripts from three inputs: (i) \textbf{Reference Code Templates} defining trajectory generation logic, including modular execution stages and inverse kinematics (IK)-based end-effector control; (ii) \textbf{High-level Task Descriptions} specifying task objectives and object semantics; and (iii) \textbf{Object Metadata} automatically retrieved from Isaac Sim~\cite{isaacsim2023}, including object poses and physical properties. Using this framework, we collect 100-150 trajectories for each task, with the volume adjusted according to task complexity and object count.

\begin{figure}[htbp]
  \centering
  \includegraphics[width=\linewidth]{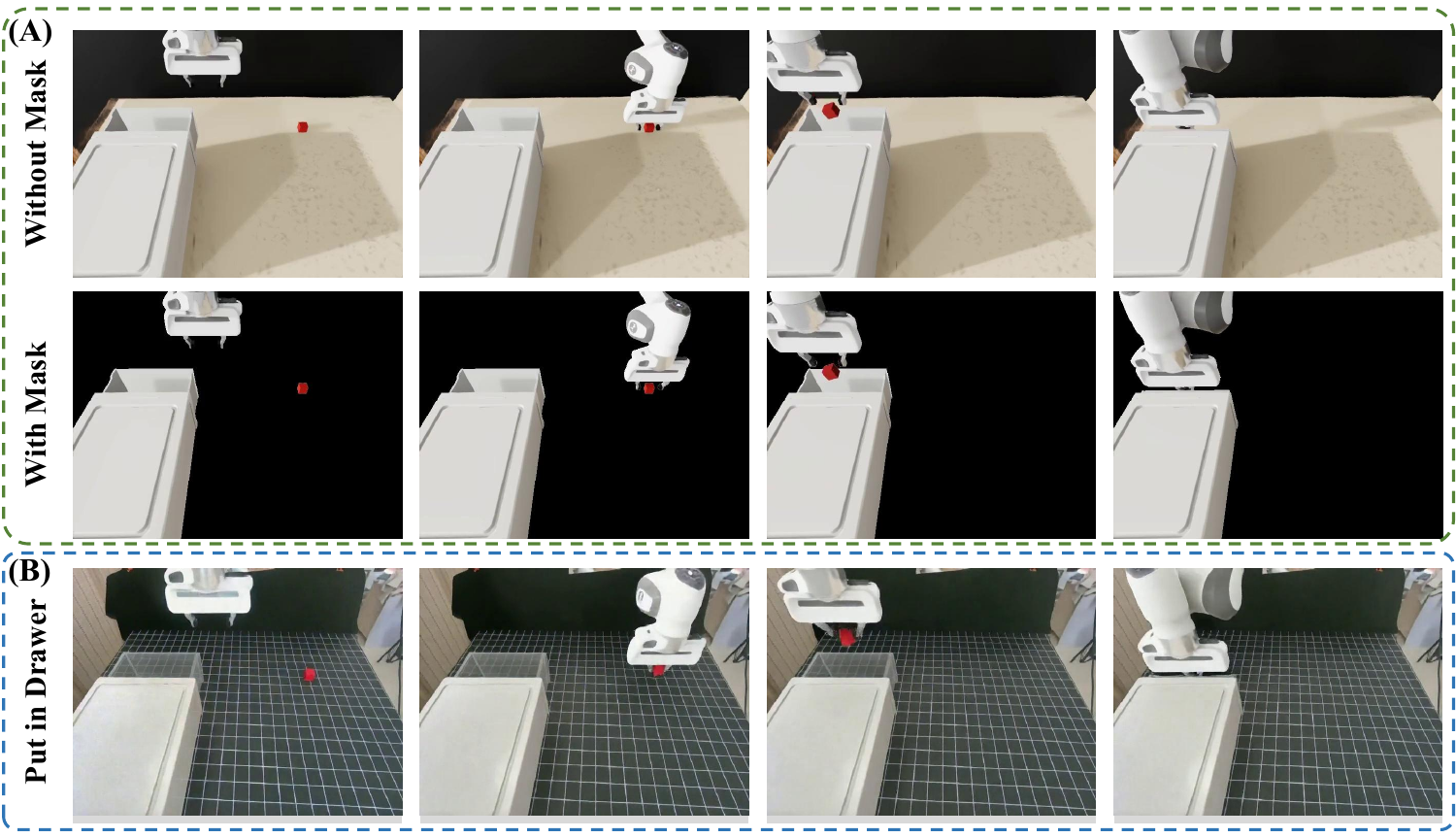}
  \caption{Qualitative results of (A) simulation data acquisition and interaction-aware masking, and (B) Grounded-Dreamer video generation.}
  \label{fig:vis}

\end{figure}

To reduce the sim-to-real domain gap, we propose an \textbf{Interaction-Aware Masking} strategy that preserves only the robotic arm and interacting objects in simulation frames (Figure~\ref{fig:vis} and \ref{fig:supp_sim}). This preprocessing encourages Cosmos-Predict 2.5~\cite{agarwal2025cosmos} and IDM~\cite{jang2025dreamgen,baker2022video} to focus on modeling interaction dynamics during fine-tuning. As a result, when conditioned on real-world initial frames, the fine-tuned Cosmos model generates realistic robot-object interactions while maintaining stable task-irrelevant backgrounds through its generative priors.

\subsection{Initial Frame Synthesis}
\label{subsec:initial_frames}

We can leverage the fine-tuned video world model to generate new rollouts conditioned on task-specific initial frames. To increase data diversity, we randomize object poses in the simulated workspace and render corresponding RGB images, instance masks, and object coordinates. We then construct preliminary scenes by compositing masked simulated objects onto real-world background images. To bridge the visual domain gap, Nano Banana 2 leverages reference photographs of the real objects to transform the composites into photorealistic images~\cite{google2026nanobanana2}. We further exploit its generative capability to augment the initial frames with variations in lighting conditions and tablecloth textures, improving robustness to distribution shifts. Additional implementation details and visualizations are provided in Appendix~\ref{subsec:details_initial_frame}.

\subsection{Data Synthesis and Closed-Loop Verification}
\label{subsec:verification}

Conditioned on the augmented initial frames, the fine-tuned Cosmos-Predict 2.5 model generates rollout videos (Figure~\ref{fig:vis} \& \ref{fig:supp_wm}). We apply Interaction-Aware Masking using SAM~\cite{kirillov2023segment} and employ the IDM to extract action trajectories, forming paired synthetic training data. 

However, generative world models may produce \textbf{physical hallucinations}, including (1) \textbf{Morphological Hallucinations}, such as invalid robot kinematics or physically infeasible joint configurations; (2) \textbf{Spatial-Depth Ambiguities}, where apparent 2D contacts do not reflect valid 3D interactions.

\begin{figure}[t]
  \centering
  \includegraphics[width=\linewidth]{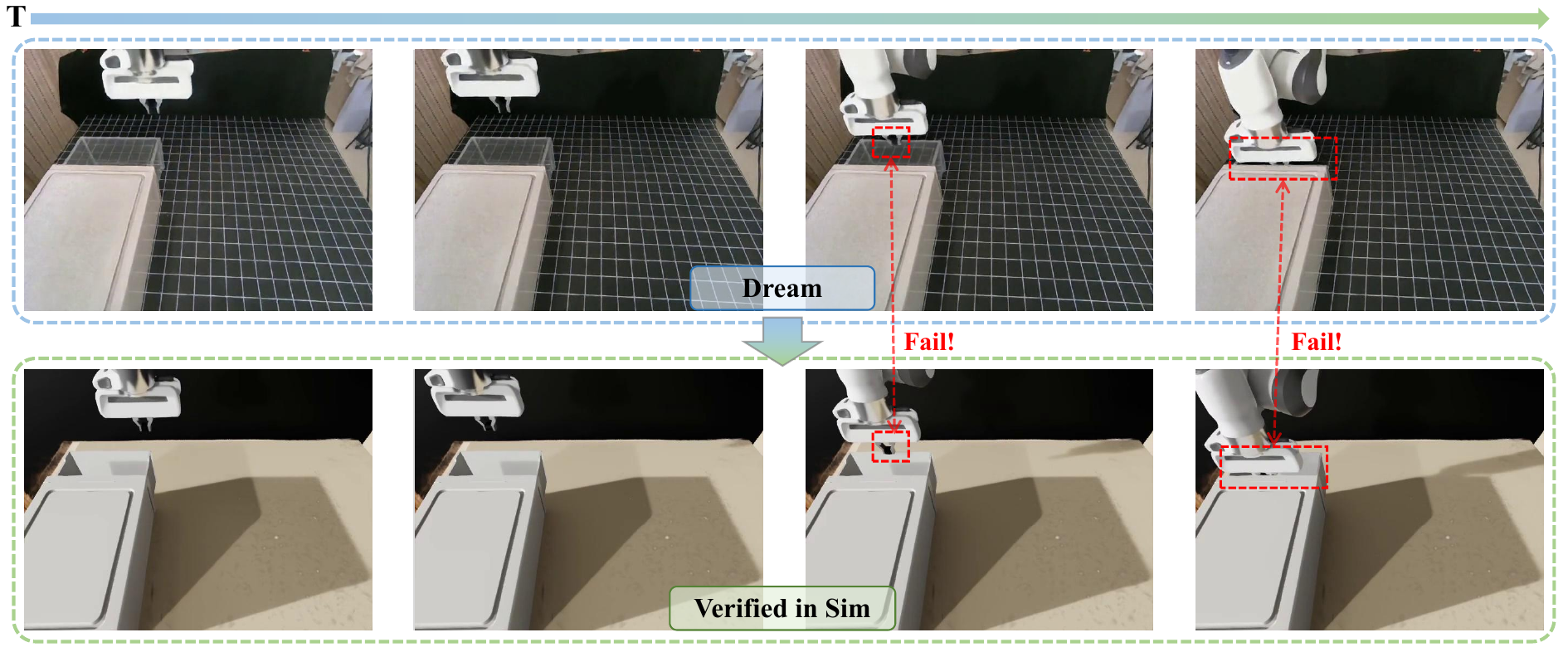}
  \caption{Visualization of Simulation Validation. The physical simulation correctly captures a task failure due to incorrect gripper placement, whereas the world model exhibits spatial hallucination by generating a hallucinated successful outcome.}
  \label{fig:varified}
  
\end{figure}

To address these issues, we introduce a \textbf{Closed-Loop Verification} mechanism. Using the reconstructed simulation environment and initialized object poses (Sec.~\ref{subsec:initial_frames}), we replay synthesized action trajectories in the physics simulator and classify them as \textit{Optimal} or \textit{Sub-optimal} according to task success criteria. Figure~\ref{fig:varified} illustrates this process. This verification stage filters out physically inconsistent or task-failing trajectories, ensuring high-quality data for downstream policy learning. Experiments in Sec.~\ref{sec:experiments} show that this filtering is critical for maintaining strong policy performance.

\begin{figure}[t]
  \centering
  \includegraphics[width=\linewidth]{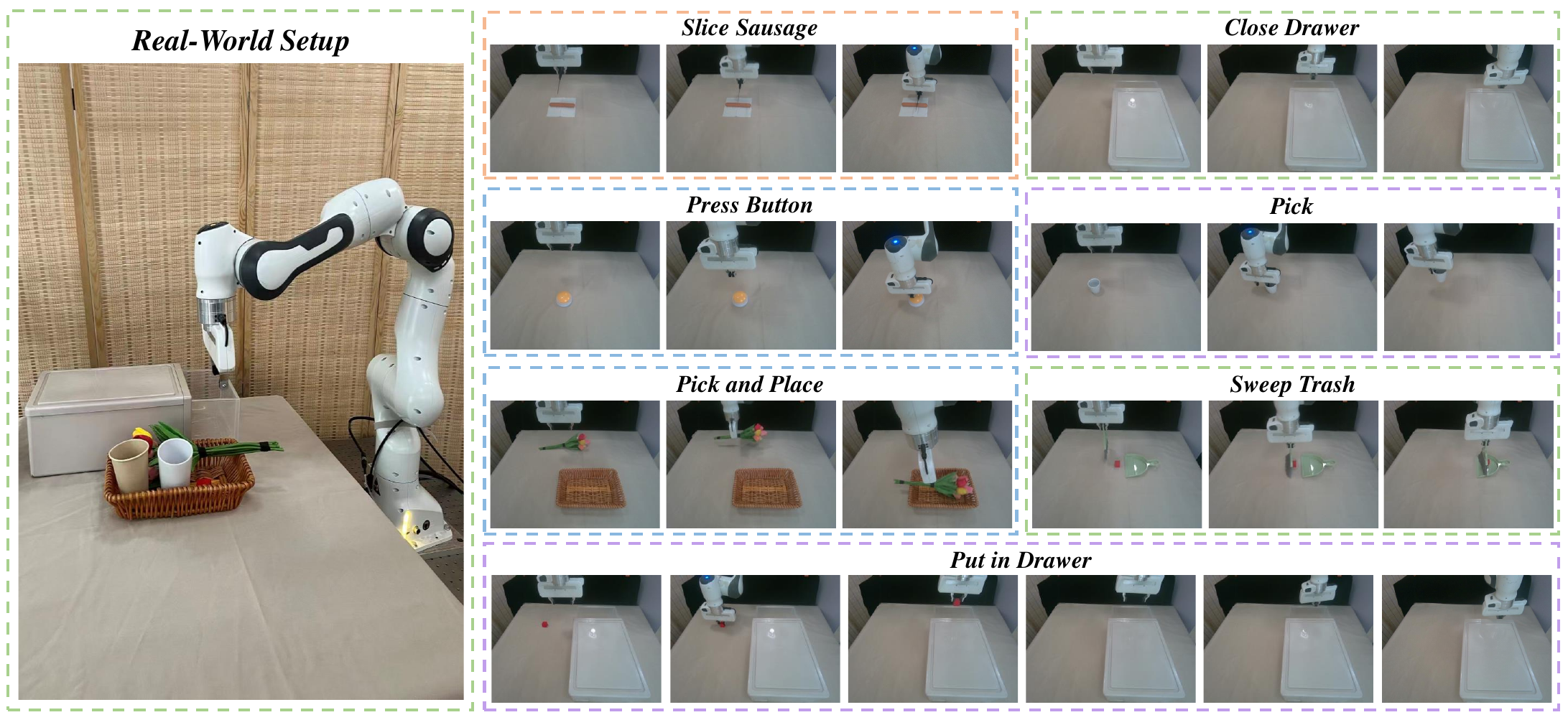}
  \caption{\textbf{Real-world Setup and Task Execution.} (Left) Real-robot evaluation environment; (Right) Visualizations of our policy executing manipulation tasks.}
  \label{fig:real}
  
\end{figure}

%% file: sections/4_exps.tex
\section{Experiments}
\label{sec:experiments}

\subsection{Experimental Setup}

\paragraph{Manipulation Tasks.} We evaluate Grounded-Dreamer across seven diverse real-world tasks: \textit{Pick}, \textit{Close Drawer}, \textit{Press Button}, \textit{Pick and Place}, \textit{Put in Drawer}, \textit{Sweep Trash}, and \textit{Slice Sausage}. This suite covers a broad spectrum of requirements, ranging from short-horizon object interactions to sophisticated tool usage and handling of deformable objects. Crucially, our evaluation incorporates both in-distribution benchmarks and challenging Out-of-Distribution scenarios, providing a rigorous assessment of the framework's zero-shot generalization across novel environments and unseen object configurations. 

\paragraph{Hardware Setup.} The physical experimental platform centers on a Franka Emika robot arm paired with a Panda gripper. Visual observations are captured by an Intel RealSense D405 depth camera. For real-time inference and control, the trained policies are deployed on a workstation equipped with an NVIDIA GeForce RTX 4090 GPU. Figure~\ref{fig:real} illustrates the comprehensive hardware configuration alongside representative real-world execution trajectories.

\newcolumntype{Y}{>{\centering\arraybackslash}X}
\begin{table}[t]
  \centering
  \small
  \setlength{\tabcolsep}{2pt} 
  \caption{\textbf{Main experimental results across seven tasks.} We compare the success rates of DP and $\pi_{0.5}$ under different data regimes, with RIGVid as a representative baseline.}
  \label{tab:main_results}
  \input{tables/table_1_}

\end{table}

\begin{table}[t]
  \caption{\textbf{Success Rates in OOD Scenarios.} Performance comparison of models trained on two types of data across seven manipulation tasks under out-of-distribution (OOD) conditions.}
  \centering
  \small
  \setlength{\tabcolsep}{2pt}
  \label{tab:ood}
  \input{tables/table_2}

\end{table}

\paragraph{Baselines and Metrics.} We employ Diffusion Policy (DP)~\cite{chi2023diffusion} and $\pi_{0.5}$~\cite{pi2025pi05} as representative policy backbones. To provide a comprehensive evaluation of Grounded-Dreamer, we compare our approach against several data regimes and alternative methodologies: 
(i) \textbf{Sim-only}: Policies trained exclusively on raw simulation data, representing a performance lower bound; 
(ii) \textbf{Real-only}: Policies trained on a full set of 50 human-labeled real-world trajectories per task; 
(iii) \textbf{Low-data Real}: A regime where only 5 real-world samples are available, representing extreme data scarcity; 
(iv) \textbf{RIGVid}~\cite{patel2025rigvid}: A baseline that utilizes a video world model to predict object-centric hand-object interactions, from which end-effector poses are extracted and mapped to robot actions.

The primary evaluation metric is the \textbf{Success Rate}. We perform 30 trials per task for scaling law analysis and 20 trials for all other comparative evaluations.

\subsection{Main Results and Comparative Analysis}
We evaluate the performance of DP and $\pi_{0.5}$ models trained with different data regimes (50 trajectories per task). As summarized in Table ~\ref{tab:main_results}, the \textbf{Sim-only} baseline performs poorly on real-world tasks, failing to complete most interactions due to significant visual and physical domain gaps. In contrast, the \textit{Real-only} baseline maintains a high success rate. Notably, policies trained exclusively on our \textbf{verified GD data} achieve competitive performance relative to the \textit{Real-only} baseline. Specifically, for the \textit{Pressing Button} task, the $\pi_{0.5}$ policy reaches parity with the performance of its real-world counterpart. This demonstrates the immense potential of world model data, particularly for labor-intensive tasks like \textit{Sweeping Trash} and \textit{Slicing Sausage}, where synthetic data can be generated autonomously from few frames at near-zero cost. We also compare our framework with RIGVid, which utilizes world models through an alternative paradigm. RIGVid performs significantly below expectations, exhibiting a substantial performance gap compared to both the Real-only and our verified GD-data regimes. This performance degradation is primarily attributed to inaccuracies in hand-pose extraction and susceptibility to viewpoint jitter.

Furthermore, we investigate Co-training using a minimal set of real demonstrations combined with 50 GD trajectories. While \textit{Low-data Real} models fail to generalize across backbones---yielding only 8.6\% for DP and 14.3\% for $\pi_{0.5}$---adding GD data leads to a substantial performance leap. This regime not only outperforms the \textit{GD-data-only} model but also matches or exceeds the 50-sample \textit{Real-only} baseline. Here, the few real trajectories act as \textbf{domain anchors}, aligning the synthetic distribution with the real domain and further unlocking the world model's latent potential.

\subsection{Generalization and Robustness to Environment Variation}
To enhance environmental diversity, we utilize the \textit{Nano Banana 2} model~\cite{google2026nanobanana2} to augment real-world initial frames with varying backgrounds, tablecloth textures, and lighting conditions before generating 50 GD trajectories. We then evaluate the DP and $\pi_{0.5}$ models in \textbf{Out-of-Distribution (OOD)} scenarios. As shown in Table~\ref{tab:ood}, models trained with GD data exhibit superior generalization capabilities, maintaining robust performance in novel environments. Conversely, the \textit{Real-only} baseline, constrained by the limited diversity of its training set, fails to generalize; its average success rates drop to only \textbf{7.1\% and 16.4\% for DP and $\pi_{0.5}$, respectively.} Crucially, this environmental variety is achieved autonomously through image-to-image augmentation of initial frames, requiring no additional physical intervention. Attaining a comparable level of diversity through manual real-world collection would necessitate exhaustive and repetitive re-collection cycles, incurring prohibitive overhead in terms of human labor and physical resources.

\begin{figure}[htbp]
  \centering
  \includegraphics[width=\linewidth]{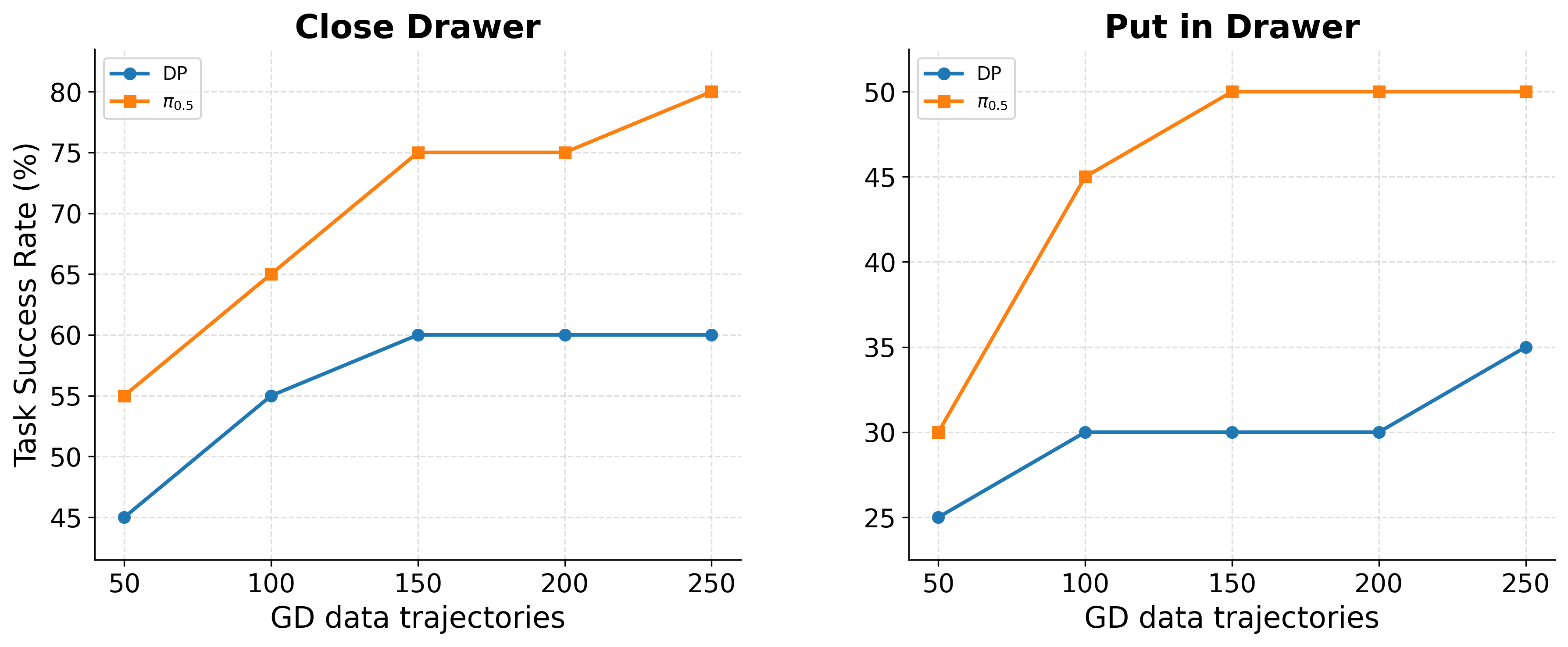}
  \caption{\textbf{Scaling Success Rates.} Success rates on two representative tasks improve as the number of Grounded-Dreamer (GD) generated trajectories increases.}
  \label{fig:scaling}
  
\end{figure}

\subsection{Scaling Laws of Synthetic World Model Data}
To evaluate the efficiency and potential ceiling of training with synthetic data, we investigate the scaling laws of GD-generated trajectories in the \textit{Close Drawer} and \textit{Put in Drawer} tasks. As illustrated in Figure~\ref{fig:scaling}, increasing the number of synthetic trajectories improves success rates for both tasks. However, performance begins to plateau at approximately 150 trajectories, with further increases yielding only marginal improvements.

\begin{table}[htbp]
  \caption{Ablation on Interaction-Aware Masking.}
  \label{tables/ablation1}
  \centering
  \small
  \setlength{\tabcolsep}{2pt}
  \input{tables/table_3}
\end{table}

\begin{table}[htbp]
  \caption{Ablation on Replay Verification.}
  \label{tables/ablation2}
  \centering
  \small
  \setlength{\tabcolsep}{2pt}
  \input{tables/table_4}

\end{table}

\subsection{Ablation Studies: Key Components of Grounded-Dreamer}
\label{subsec:ablations}

In this section, we evaluate the individual contributions of two critical components of our framework: \textbf{Interaction-Aware Masking} and \textbf{Closed-Loop Replay Verification}.

\begin{figure}[!h]
  \centering
  \includegraphics[width=\linewidth]{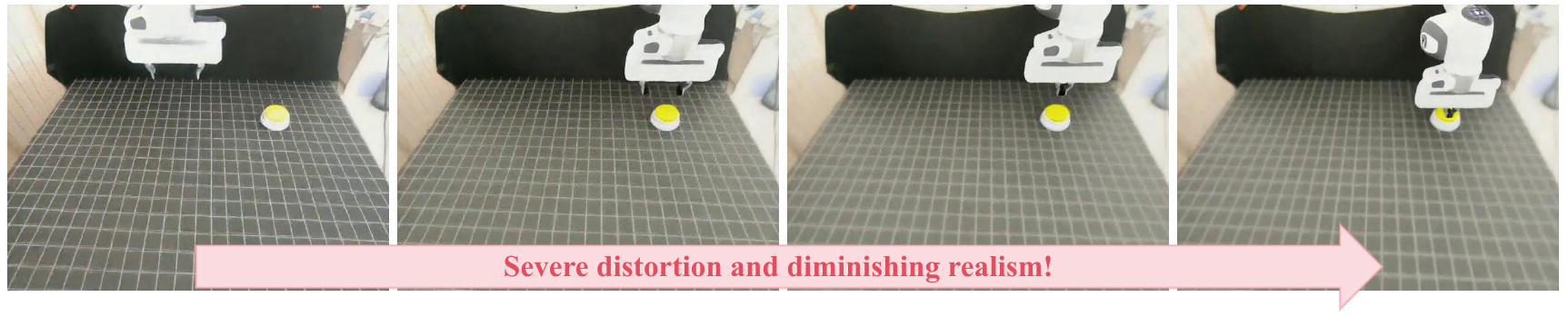}
  \caption{\textbf{Ablation without Masking.} Visualizations of video rollouts generated by the world model fine-tuned on raw simulation data without the interaction-aware masking strategy.}
  \label{fig:xxx}
  
\end{figure}

\paragraph{Effect of Interaction-Aware Masking.} We compare downstream policies trained on synthetic data from world models fine-tuned with either interaction-aware masks or raw simulation data. Without masking, generated videos exhibit distorted pixels and unrealistic scene deformations (Fig.~\ref{fig:xxx}), and success rates drop significantly (Table~\ref{tables/ablation1}). These results support masking task-irrelevant regions during world-model fine-tuning.

\paragraph{Necessity of Closed-Loop Replay Verification.} Table~\ref{tables/ablation2} compares policies trained with varying ratios of verified (\textit{Optimal}) versus non-verified (\textit{Sub-optimal}) trajectories. Task success increases monotonically with the proportion of verified data. Notably, at 50\% sub-optimal data, performance \textbf{suffers a precipitous drop}. These results confirm the importance of simulation replay for filtering synthetic training data.

%% file: tables/table_1_.tex
\begin{tabularx}{\textwidth}{lYYYYYYY}
    \toprule
    \textbf{Method} & Pick & Button & Close Drawer & Pick and Place & Sweep Trash & Slice Sausage & Put in Drawer \\
    \midrule
    Sim-only (DP)        & 10\% & 10\% & 0\%  & 5\%  & 5\%  & 0\%  & 0\%  \\
    Real-only (DP)       & \underline{50\%} & \underline{60\%} & \textbf{60\%} & \textbf{50\%} & \underline{50\%} & \textbf{70\%} & \underline{35\%} \\
    GD data (DP)         & \textit{45\%} & \textit{50\%} & \textit{45\%} & \textit{35\%} & \textit{45\%} & \textit{55\%} & \textit{25\%} \\
    Low-data Real (DP)   & 10\% & 10\% & 15\% & 0\%  & 10\% & 10\% & 5\%  \\
    Co-training (DP)     & \textbf{55\%} & \textbf{65\%} & \underline{50\%} & \underline{45\%} & \textbf{55\%} & \underline{60\%} & \textbf{40\%} \\
    \midrule
    Sim-only ($\pi_{0.5}$) & 10\% & 15\% & \textit{15\%} & 5\%  & 10\% & 5\%  & 0\%  \\
    Real-only ($\pi_{0.5}$) & \underline{65\%} & \underline{75\%} & \textbf{65\%} & \textbf{60\%} & \underline{60\%} & \textbf{75\%} & \textbf{45\%} \\
    GD data ($\pi_{0.5}$)   & \textit{55\%} & \underline{75\%} & \underline{55\%} & \textit{50\%} & \underline{60\%} & \textit{60\%} & \underline{30\%} \\
    Low-data Real ($\pi_{0.5}$) & 20\% & 20\% & 10\% & 15\% & 20\% & 10\% & 5\%  \\
    Co-training ($\pi_{0.5}$) & \textbf{70\%} & \textbf{80\%} & \textbf{65\%} & \underline{55\%} & \textbf{75\%} & \underline{70\%} & \textbf{45\%} \\
    \midrule
    RIGVid               & 25\% & 20\% & 5\%  & 5\%  & 0\%  & 0\%  & 0\%  \\
    \bottomrule
\end{tabularx}

%% file: tables/table_2.tex
\begin{tabularx}{\textwidth}{lYYYYYYY}
    \toprule
    \textbf{Method} & Pick & Button & Close Drawer & Pick and Place & Sweep Trash & Slice Sausage & Put in Drawer \\
    \midrule
    Real-only (DP)       
        & 10\% 
        & 5\%  
        & 5\%  
        & 0\%  
        & 5\%  
        & 5\%  
        & 0\%  \\
    Verified WM data (DP) 
        & \textbf{40\%} 
        & \textbf{50\%} 
        & \textbf{35\%} 
        & \textbf{30\%} 
        & \textbf{40\%} 
        & \textbf{45\%} 
        & \textbf{15\%} \\
    \midrule
    Real-only ($\pi_{0.5}$)       
        & 25\% 
        & 20\% 
        & 10\% 
        & 20\% 
        & 20\% 
        & 15\% 
        & 5\%  \\
    Verified WM data ($\pi_{0.5}$) 
        & \textbf{45\%} 
        & \textbf{60\%} 
        & \textbf{55\%} 
        & \textbf{40\%} 
        & \textbf{60\%} 
        & \textbf{50\%} 
        & \textbf{30\%} \\
    \bottomrule
\end{tabularx}

%% file: tables/table_3.tex
\begin{tabularx}{\textwidth}{lYYYYYYY}
    \toprule
    \textbf{Method} & Pick & Button & Close Drawer & Pick and Place & Sweep Trash & Slice Sausage & Put in Drawer \\
    \midrule
    w/ Masking (DP)    
        & \textbf{45\%} 
        & \textbf{50\%} 
        & \textbf{45\%} 
        & \textbf{35\%} 
        & \textbf{45\%} 
        & \textbf{55\%} 
        & \textbf{25\%} \\
    w/o Masking (DP)   
        & 0\%  
        & 5\%  
        & 0\%  
        & 0\%  
        & 0\%  
        & 0\%  
        & 0\%  \\
    \midrule
    w/ Masking ($\pi_{0.5}$)  
        & \textbf{55\%} 
        & \textbf{75\%} 
        & \textbf{55\%} 
        & \textbf{50\%} 
        & \textbf{60\%} 
        & \textbf{60\%} 
        & \textbf{30\%} \\
    w/o Masking ($\pi_{0.5}$) 
        & 5\%  
        & 0\%  
        & 10\% 
        & 0\%  
        & 10\% 
        & 5\%  
        & 0\%  \\
    \bottomrule
\end{tabularx}

%% file: tables/table_4.tex
\begin{tabularx}{\textwidth}{lYYYYYYY}
    \toprule
    \textbf{Method} & Pick & Button & Close Drawer & Pick and Place & Sweep Trash & Slice Sausage & Put in Drawer \\
    \midrule
    100\% (DP) & \textbf{45\%} & \textbf{50\%} & \textbf{45\%} & \textbf{35\%} & \textbf{45\%} & \textbf{55\%} & \textbf{25\%} \\
    80\% (DP)  & 40\% & 40\% & 35\% & 30\% & 30\% & 45\% & 10\% \\
    50\% (DP)  & 25\% & 30\% & 25\% & 25\% & 20\% & 20\% & 5\%  \\
    \midrule
    100\% ($\pi_{0.5}$) & \textbf{55\%} & \textbf{75\%} & \textbf{55\%} & \textbf{50\%} & \textbf{60\%} & \textbf{60\%} & \textbf{30\%} \\
    80\% ($\pi_{0.5}$)  & 35\% & 60\% & 50\% & 35\% & 45\% & 50\% & 20\% \\
    50\% ($\pi_{0.5}$)  & 25\% & 35\% & 25\% & 20\% & 25\% & 30\% & 10\% \\
    \bottomrule
  \end{tabularx}

%% file: supp.tex
\input{supp/A}

\input{supp/B-}
\input{supp/B}
\input{supp/C}
\input{supp/D}

%% file: supp/A.tex
\section{Limitation}
\label{limitation}

While Grounded-Dreamer substantially automates data generation, it currently requires a minimal, one-time collection of reference images---typically one scene overview and 1--4 object perspectives---to initialize the environment. Given that image-conditioned editing models are still rapidly evolving, the success rate for our synthesized initial frames is approximately 65\%. However, we view this as a robust baseline that will naturally improve as foundational generative models mature. 

While our experiments establish the framework's effectiveness with hundreds of trajectories per task, the scale was primarily constrained by available computational resources rather than algorithmic bottlenecks. The inherent design of our pipeline is highly scalable; as hardware constraints are alleviated, we anticipate extending this paradigm to massive-scale data generation to further unlock the scaling potential of generalist robot policies.

%% file: supp/B-.tex
\section{Broader Impacts}

Our work introduces a scalable pipeline for generating high-fidelity, verified synthetic data, which has significant implications for the future of Embodied AI. By reducing the reliance on costly and labor-intensive real-world data collection, our approach facilitates the scaling of robotic learning across diverse tasks and environments. While this may accelerate the deployment of autonomous systems, it also underscores the importance of addressing potential biases in world models and ensuring the safety of policies trained on synthetic data. We believe this research promotes more accessible and efficient robot learning, though we encourage practitioners to maintain rigorous safety protocols when deploying such models in real-world human-centric environments.

%% file: supp/B.tex
\section{Additional Qualitative Visualizations}
\label{appendix:visualization}
\begin{figure}[htbp]
  \centering
  \includegraphics[width=\linewidth]{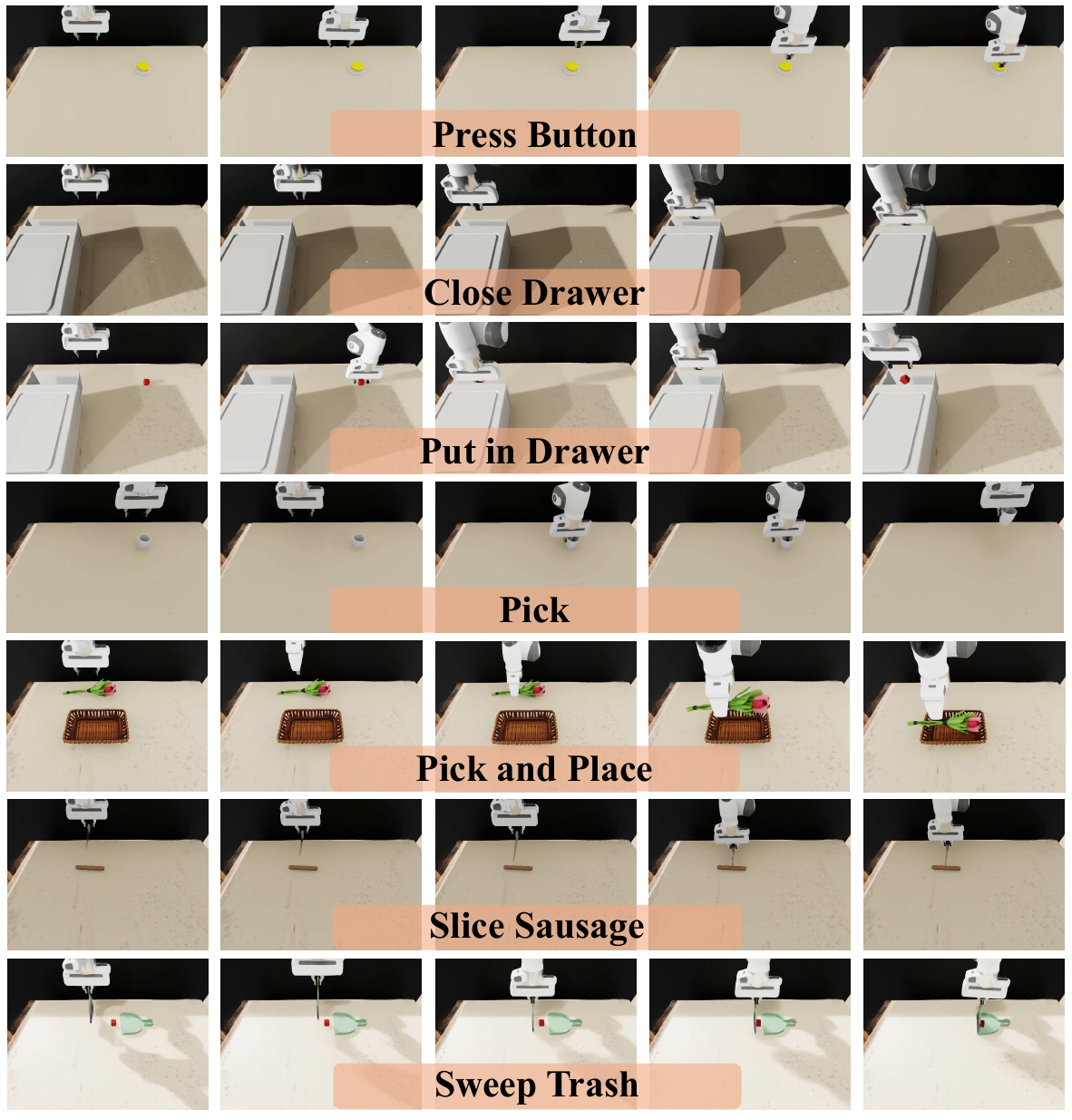}
  \caption{Additional visualizations of automated data collection in simulation.}
  \label{fig:supp_sim}
\end{figure}
In this section, we provide extended visualization results to facilitate a comprehensive qualitative analysis of the \textit{Grounded-Dreamer} framework.

\subsection{Automated Simulation-Based Data Acquisition}
\label{subsec:sim_data_collection}

Figure~\ref{fig:supp_sim} illustrates the automated data collection process within the simulation environment. For each task, the top row displays the original simulation frames, while the bottom row shows the frames after applying our interaction-aware masking strategy.

\subsection{Video Rollouts from Grounded-Dreamer}
\label{subsec:dreamer_rollouts}

\begin{figure}[htbp]
  \centering
  \includegraphics[width=\linewidth]{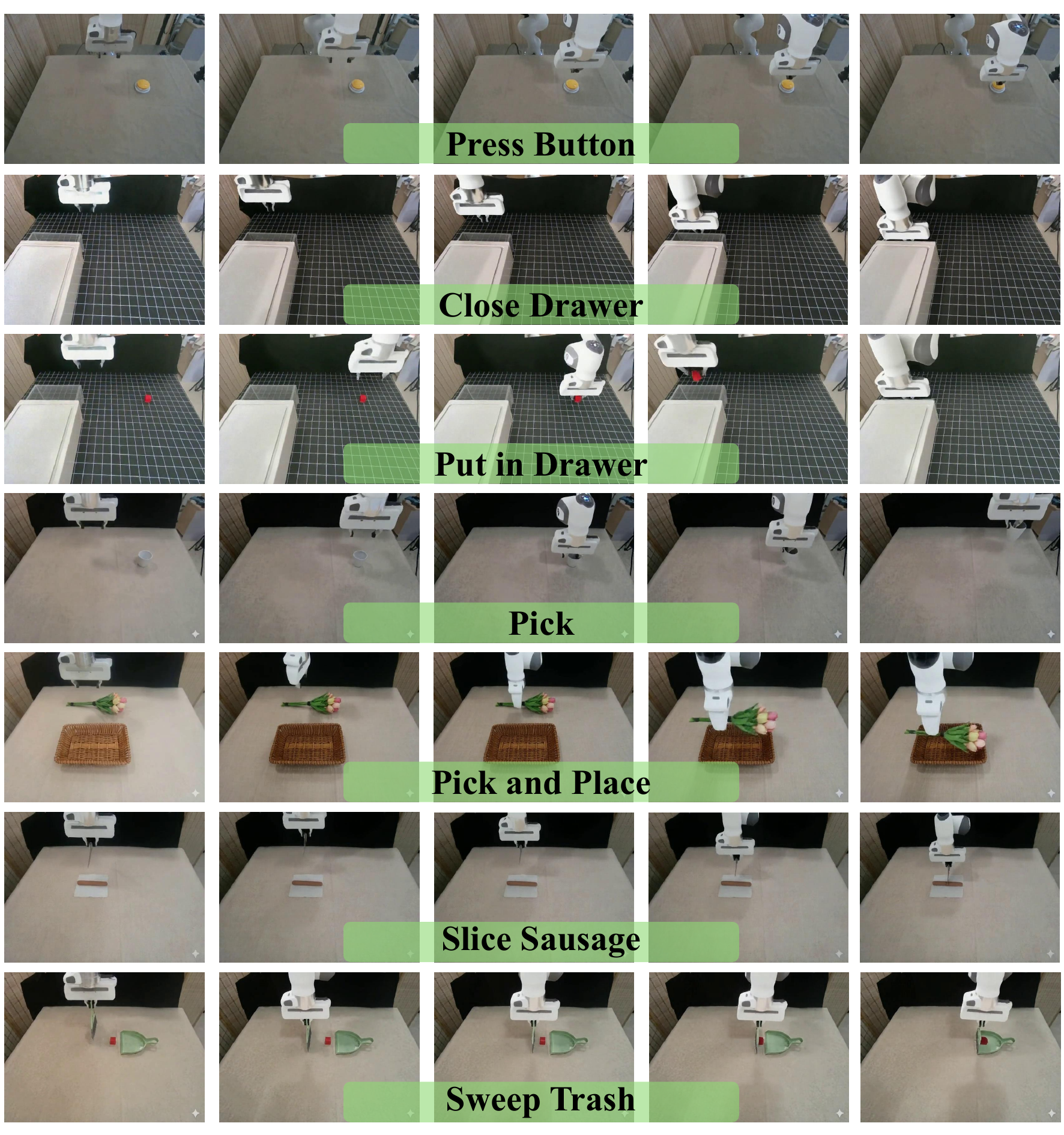}
  \caption{Additional visualizations of Grounded-Dreamer video generation.}
  \label{fig:supp_wm}
\end{figure}
We present additional video rollout results generated by our world model in Figure~\ref{fig:supp_wm}. These sequences demonstrate the \textit{Grounded-Dreamer}'s capability to predict physically plausible future states. 

%% file: supp/C.tex
\section{Details of Initial Frame Synthesis}
\label{subsec:details_initial_frame}

We employ \textit{Nano Banana 2}~\cite{google2026nanobanana2} as the generative backbone for synthesizing initial frames that are distributionally aligned with the real world. The implementation details are as follows:

\begin{itemize}
    \item \textbf{Input Modalities:} The synthesis process takes two distinct inputs: (1) a composite image consisting of simulated object assets overlaid on a real-world scene, and (2) a reference photograph containing the actual objects to provide essential structural and illumination constraints.
    \item \textbf{Prompting Technique:} We utilize a structured prompt to guide the generative harmonization: ``There is a simulated object in Image 1, identified as [object]. Image 2 displays its true, real-life form. Please substitute the simulated object in Image 1 with a new version in a realistic style, while keeping the position and orientation of the original simulated object from Image 1. The replacement should not be aligned with the realistic object in Image 2. The real object serves only as a visual guide for appearance. The background must not be altered.''
\end{itemize}

\begin{figure}[htbp]
  \centering
  \includegraphics[width=\linewidth]{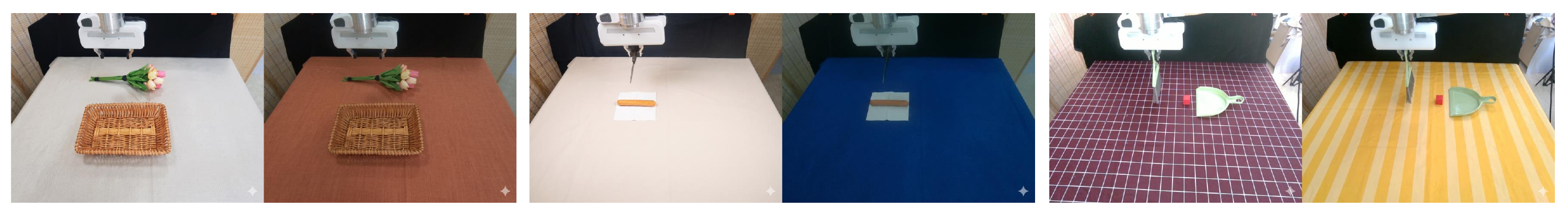}
  \caption{Qualitative Visualizations of Refined Initial Frames.}
  \label{fig:supp_init}
\end{figure}

Figure~\ref{fig:supp_init} provides qualitative visualizations of these synthesized frames, illustrating the seamless integration of virtual assets into diverse physical contexts.

%% file: supp/D.tex
\section{Experimental Details}
\label{appendix:experimental_details}

In this section, we provide comprehensive details regarding the training and inference pipelines of our framework to ensure reproducibility.

\subsection{Fine-Tuning Details of Cosmos}
\label{subsec:cosmos_details}
We fine-tune the 2B-parameter Cosmos Predict2.5 Video2World model \cite{agarwal2025cosmos} as the generative backbone. The model is optimized using a learning rate of $4.315 \times 10^{-5}$ on a cluster of eight NVIDIA A800 GPUs with context parallelism. Training is conducted for 1,400 to 2,000 iterations with a global batch size of 32. We use a temporal length of 125 frames for the \textit{Put in Drawer} task and 93 frames for all other tasks. The fine-tuning process requires 74 GB of VRAM per GPU.
\subsection{Training Details of the Inverse Dynamics Model (IDM)}
\label{subsec:idm_details}
The IDM \cite{jang2025dreamgen} features a SigLIP2-Large encoder and a flow-matching action head. It is trained across the eight A800 GPUs for 5,000 to 10,000 steps with a global batch size of 512 and a $10^{-4}$ learning rate. Training is performed in BF16 mixed precision to ensure numerical stability.
\subsection{Synthetic Data Generation and Action Extraction}
\label{subsec:generation_details}
All inference and data generation tasks are also executed on the eight NVIDIA A800 GPU cluster.\textbf{Video Generation via Cosmos.} We generate rollout sequences conditioned on five replicated initial frames. For the \textit{Put in Drawer} task, the model produces 125-frame sequences, while 93-frame sequences are generated for other tasks. The inference memory footprint is approximately 16 GB per GPU.\textbf{Action Extraction via IDM.} Robot actions are extracted from the generated sequences using the IDM, which demonstrates high throughput (0.5 seconds per trajectory). This process is memory-efficient, requiring only 4 GB of VRAM per GPU. The average task success rate during this generation phase is approximately 45\%.

%% file: main.bbl
\begin{thebibliography}{48}
\providecommand{\natexlab}[1]{#1}
\providecommand{\url}[1]{\texttt{#1}}
\expandafter\ifx\csname urlstyle\endcsname\relax
  \providecommand{\doi}[1]{doi: #1}\else
  \providecommand{\doi}{doi: \begingroup \urlstyle{rm}\Url}\fi

\bibitem[Agrawal et~al.(2016)Agrawal, Nair, Abbeel, Malik, and
  Levine]{agrawal2016learning}
Pulkit Agrawal, Ashvin Nair, Pieter Abbeel, Jitendra Malik, and Sergey Levine.
\newblock Learning to poke by poking: Experiential learning of intuitive
  physics.
\newblock In \emph{Advances in Neural Information Processing Systems}, 2016.

\bibitem[Baker et~al.(2022)Baker, Akkaya, Zhokhov, Huizinga, Tang, Ecoffet,
  Houghton, Sampedro, and Clune]{baker2022video}
Bowen Baker, Ilge Akkaya, Peter Zhokhov, Joost Huizinga, Jie Tang, Adrien
  Ecoffet, Brandon Houghton, Raul Sampedro, and Jeff Clune.
\newblock Video pretraining ({VPT}): Learning to act by watching unlabeled
  online videos.
\newblock \emph{Advances in Neural Information Processing Systems},
  35:\penalty0 24639--24654, 2022.

\bibitem[Black et~al.(2024)Black, Brown, Driess, Esmail, Equi, Finn, Fusai,
  Groom, Hausman, Ichter, Jakubczak, Jones, Ke, Levine, Li-Bell, Mothukuri,
  Nair, Pertsch, Shi, Tanner, Vuong, Walling, Wang, and
  Zhilinsky]{black2024pi0}
Kevin Black, Noah Brown, Danny Driess, Adnan Esmail, Michael Equi, Chelsea
  Finn, Niccolo Fusai, Lachy Groom, Karol Hausman, Brian Ichter, Szymon
  Jakubczak, Tim Jones, Liyiming Ke, Sergey Levine, Adrian Li-Bell, Mohith
  Mothukuri, Suraj Nair, Karl Pertsch, Lucy~Xiaoyang Shi, James Tanner, Quan
  Vuong, Anna Walling, Haohuan Wang, and Ury Zhilinsky.
\newblock {{$\pi_0$}}: A {Vision-Language-Action} flow model for general robot
  control.
\newblock \emph{arXiv preprint arXiv:2410.24164}, 2024.

\bibitem[Bruce et~al.(2024)Bruce, Dennis, Edwards, Parker-Holder, Shi, Hughes,
  Lai, Mavalankar, Steigerwald, Apps, Aytar, Bechtle, Behbahani, Chan, Heess,
  Gonzalez, Osindero, Ozair, Reed, Zhang, Zolna, Clune, de~Freitas, and
  Rockt{\"a}schel]{bruce2024genie}
Jake Bruce, Michael~D. Dennis, Ashley Edwards, Jack Parker-Holder, Yuge Shi,
  Edward Hughes, Matthew Lai, Aditi Mavalankar, Ria Steigerwald, Chris Apps,
  Yusuf Aytar, Sarah Bechtle, Feryal Behbahani, Stephanie C.~Y. Chan, Nicolas
  Heess, Luis Gonzalez, Simon Osindero, Sherjil Ozair, Scott Reed, Jingwei
  Zhang, Konrad Zolna, Jeff Clune, Nando de~Freitas, and Tim Rockt{\"a}schel.
\newblock {{Genie}}: Generative interactive environments.
\newblock In \emph{International Conference on Machine Learning}, 2024.

\bibitem[Chi et~al.(2024)Chi, Xu, Pan, Cousineau, Burchfiel, Feng, Tedrake, and
  Song]{chi2024universal}
Cheng Chi, Zhenjia Xu, Chuer Pan, Eric Cousineau, Benjamin Burchfiel, Siyuan
  Feng, Russ Tedrake, and Shuran Song.
\newblock Universal manipulation interface: In-the-wild robot teaching without
  in-the-wild robots.
\newblock \emph{arXiv preprint arXiv:2402.10329}, 2024.

\bibitem[Chi et~al.(2025)Chi, Xu, Feng, Cousineau, Du, Burchfiel, Tedrake, and
  Song]{chi2023diffusion}
Cheng Chi, Zhenjia Xu, Siyuan Feng, Eric Cousineau, Yilun Du, Benjamin
  Burchfiel, Russ Tedrake, and Shuran Song.
\newblock {{Diffusion Policy}}: Visuomotor policy learning via action
  diffusion.
\newblock \emph{The International Journal of Robotics Research}, 44\penalty0
  (10--11):\penalty0 1684--1704, 2025.
\newblock \doi{10.1177/02783649241273668}.

\bibitem[Finn \& Levine(2017)Finn and Levine]{finn2017visualforesight}
Chelsea Finn and Sergey Levine.
\newblock Deep visual foresight for planning robot motion.
\newblock In \emph{IEEE International Conference on Robotics and Automation},
  pp.\  2786--2793, 2017.

\bibitem[Garrett et~al.(2025)Garrett, Mandlekar, Wen, and
  Fox]{garrett2024skillmimicgen}
Caelan Garrett, Ajay Mandlekar, Bowen Wen, and Dieter Fox.
\newblock {{SkillMimicGen}}: Automated demonstration generation for efficient
  skill learning and deployment.
\newblock In \emph{Proceedings of the 8th Conference on Robot Learning}, volume
  270 of \emph{Proceedings of Machine Learning Research}, pp.\  2750--2790,
  2025.

\bibitem[{GigaBrain Team}(2025)]{gigabrain2025}
{GigaBrain Team}.
\newblock {{GigaBrain-0}}: A world model-powered vision-language-action model.
\newblock \emph{arXiv preprint arXiv:2510.19430}, 2025.

\bibitem[{Google DeepMind}(2026)]{google2026nanobanana2}
{Google DeepMind}.
\newblock {Gemini 3.1 Flash Image} ({Nano Banana 2}): Model card.
\newblock Technical report, Google DeepMind, 2026.
\newblock URL
  \url{https://deepmind.google/models/model-cards/gemini-3-1-flash-image/}.
\newblock Updated 26 February 2026; Accessed: 2026-05-07.

\bibitem[Ha \& Schmidhuber(2018)Ha and Schmidhuber]{ha2018worldmodels}
David Ha and J{\"u}rgen Schmidhuber.
\newblock World models.
\newblock \emph{arXiv preprint arXiv:1803.10122}, 2018.

\bibitem[Hafner et~al.(2020)Hafner, Lillicrap, Ba, and
  Norouzi]{hafner2019dreamer}
Danijar Hafner, Timothy Lillicrap, Jimmy Ba, and Mohammad Norouzi.
\newblock Dream to control: Learning behaviors by latent imagination.
\newblock In \emph{International Conference on Learning Representations}, 2020.

\bibitem[Hafner et~al.(2025)Hafner, Pasukonis, Ba, and
  Lillicrap]{hafner2023dreamerv3}
Danijar Hafner, Jurgis Pasukonis, Jimmy Ba, and Timothy Lillicrap.
\newblock Mastering diverse control tasks through world models.
\newblock \emph{Nature}, 640:\penalty0 647--653, 2025.
\newblock \doi{10.1038/s41586-025-08744-2}.

\bibitem[Ho et~al.(2022)Ho, Chan, Saharia, Whang, Gao, Gritsenko, Kingma,
  Poole, Norouzi, Fleet, and Salimans]{ho2022videodiffusion}
Jonathan Ho, William Chan, Chitwan Saharia, Jay Whang, Ruiqi Gao, Alexey
  Gritsenko, Diederik~P. Kingma, Ben Poole, Mohammad Norouzi, David~J. Fleet,
  and Tim Salimans.
\newblock Video diffusion models.
\newblock In \emph{Advances in Neural Information Processing Systems}, 2022.

\bibitem[Hodan et~al.(2018)Hodan, Michel, Brachmann, Kehl, Buch, Kraft, Drost,
  Vidal, Ihrke, Zabulis, Sahin, Manhardt, Tombari, Kim, Matas, and
  Rother]{hodan2018bop}
Tomas Hodan, Frank Michel, Eric Brachmann, Wadim Kehl, Anders~Glent Buch, Dirk
  Kraft, Bertram Drost, Joel Vidal, Stephan Ihrke, Xenophon Zabulis, Caner
  Sahin, Fabian Manhardt, Federico Tombari, Tae-Kyun Kim, Jiri Matas, and
  Carsten Rother.
\newblock {{BOP}}: Benchmark for {6D} object pose estimation.
\newblock In \emph{European Conference on Computer Vision Workshops}, 2018.

\bibitem[Hoque et~al.(2024)Hoque, Mandlekar, Garrett, Goldberg, and
  Fox]{hoque2024intervengen}
Ryan Hoque, Ajay Mandlekar, Caelan Garrett, Ken Goldberg, and Dieter Fox.
\newblock {{IntervenGen}}: Interventional data generation for robust and
  data-efficient robot imitation learning.
\newblock In \emph{IEEE/RSJ International Conference on Intelligent Robots and
  Systems}, 2024.

\bibitem[Hua et~al.(2025)Hua, Liu, Macaluso, Lin, Zhang, Xu, and
  Wang]{hua2024gensim2}
Pu~Hua, Minghuan Liu, Annabella Macaluso, Yunfeng Lin, Weinan Zhang, Huazhe Xu,
  and Lirui Wang.
\newblock {{GenSim2}}: Scaling robot data generation with multi-modal and
  reasoning {LLMs}.
\newblock In \emph{Proceedings of the 8th Conference on Robot Learning}, volume
  270 of \emph{Proceedings of Machine Learning Research}, pp.\  5030--5066,
  2025.

\bibitem[James et~al.(2020)James, Ma, Arrojo, and Davison]{james2020rlbench}
Stephen James, Zicong Ma, David~Rovick Arrojo, and Andrew~J. Davison.
\newblock {{RLBench}}: The robot learning benchmark and learning environment.
\newblock \emph{IEEE Robotics and Automation Letters}, 5\penalty0 (2):\penalty0
  3019--3026, 2020.

\bibitem[Jang et~al.(2025)Jang, Ye, Lin, Xiang, Bjorck, Fang, Hu, Huang,
  Kundalia, Lin, Magne, Mandlekar, Narayan, Tan, Wang, Wang, Wang, Xu, Zeng,
  Zheng, Zheng, Liu, Zettlemoyer, Fox, Kautz, Reed, Zhu, and
  Fan]{jang2025dreamgen}
Joel Jang, Seonghyeon Ye, Zongyu Lin, Jiannan Xiang, Johan Bjorck, Yu~Fang,
  Fengyuan Hu, Spencer Huang, Kaushil Kundalia, Yen-Chen Lin, Loic Magne, Ajay
  Mandlekar, Avnish Narayan, You~Liang Tan, Guanzhi Wang, Jing Wang, Qi~Wang,
  Yinzhen Xu, Xiaohui Zeng, Kaiyuan Zheng, Ruijie Zheng, Ming-Yu Liu, Luke
  Zettlemoyer, Dieter Fox, Jan Kautz, Scott Reed, Yuke Zhu, and Linxi Fan.
\newblock {{DREAMGEN}}: Unlocking generalization in robot learning through
  neural trajectories.
\newblock In \emph{Proceedings of the 9th Conference on Robot Learning}, volume
  305 of \emph{Proceedings of Machine Learning Research}, pp.\  5170--5194,
  2025.

\bibitem[Jiang et~al.(2025)Jiang, Xie, Lin, Xu, Wan, Mandlekar, Fan, and
  Zhu]{jiang2024dexmimicgen}
Zhenyu Jiang, Yuqi Xie, Kevin Lin, Zhenjia Xu, Weikang Wan, Ajay Mandlekar,
  Linxi Fan, and Yuke Zhu.
\newblock {{DexMimicGen}}: Automated data generation for bimanual dexterous
  manipulation via imitation learning.
\newblock In \emph{IEEE International Conference on Robotics and Automation},
  2025.

\bibitem[Kerbl et~al.(2023)Kerbl, Kopanas, Leimk{\"u}hler, and
  Drettakis]{kerbl2023gaussian}
Bernhard Kerbl, Georgios Kopanas, Thomas Leimk{\"u}hler, and George Drettakis.
\newblock {{3D} Gaussian Splatting} for real-time radiance field rendering.
\newblock \emph{ACM Transactions on Graphics}, 42\penalty0 (4):\penalty0
  139:1--139:14, 2023.

\bibitem[Khazatsky et~al.(2024)Khazatsky, Pertsch, Nair, Balakrishna, Dasari,
  Karamcheti, Nasiriany, Srirama, Florence, Levine, Finn,
  et~al.]{khazatsky2024droid}
Alexander Khazatsky, Karl Pertsch, Suraj Nair, Ashwin Balakrishna, Sudeep
  Dasari, Siddharth Karamcheti, Soroush Nasiriany, Mohan~Kumar Srirama, Pete
  Florence, Sergey Levine, Chelsea Finn, et~al.
\newblock {{DROID}}: A large-scale in-the-wild robot manipulation dataset.
\newblock In \emph{Robotics: Science and Systems}, 2024.
\newblock \doi{10.15607/RSS.2024.XX.120}.

\bibitem[Kirillov et~al.(2023)Kirillov, Mintun, Ravi, Mao, Rolland, Gustafson,
  Xiao, Whitehead, Berg, Lo, Doll{\'a}r, and Girshick]{kirillov2023segment}
Alexander Kirillov, Eric Mintun, Nikhila Ravi, Hanzi Mao, Chloe Rolland, Laura
  Gustafson, Tete Xiao, Spencer Whitehead, Alexander~C. Berg, Wan-Yen Lo, Piotr
  Doll{\'a}r, and Ross Girshick.
\newblock Segment anything.
\newblock In \emph{IEEE/CVF International Conference on Computer Vision}, pp.\
  4015--4026, 2023.

\bibitem[Li et~al.(2026)Li, Yang, Xie, Xu, Chen, Wang, Shen, Shen, Chen, Li,
  Zheng, Zhang, Lin, Teng, Yang, Chen, Xie, and Wu]{li2025lehome}
Zeyi Li, Yushi Yang, Shawn Xie, Kyle Xu, Tianxing Chen, Yuran Wang, Zhenhao
  Shen, Yan Shen, Yue Chen, Wenjun Li, Yukun Zheng, Chaorui Zhang, Siyi Lin,
  Fei Teng, Hongjun Yang, Ming Chen, Steve Xie, and Ruihai Wu.
\newblock {{LeHome}}: A simulation environment for deformable object
  manipulation in household scenarios, 2026.
\newblock URL \url{https://arxiv.org/abs/2604.22363}.

\bibitem[Makoviychuk et~al.(2021)Makoviychuk, Wawrzyniak, Guo, Lu, Storey,
  Macklin, Hoeller, Rudin, Allshire, Handa, and State]{makoviychuk2021isaacgym}
Viktor Makoviychuk, Lukasz Wawrzyniak, Yunrong Guo, Michelle Lu, Kier Storey,
  Miles Macklin, David Hoeller, Nikita Rudin, Arthur Allshire, Ankur Handa, and
  Gavriel State.
\newblock {{Isaac Gym}}: High performance {GPU}-based physics simulation for
  robot learning.
\newblock In \emph{NeurIPS 2021 Datasets and Benchmarks Track}, 2021.

\bibitem[Mandlekar et~al.(2023)Mandlekar, Nasiriany, Wen, Akinola, Narang, Fan,
  Zhu, and Fox]{mandlekar2023mimicgen}
Ajay Mandlekar, Soroush Nasiriany, Bowen Wen, Iretiayo Akinola, Yashraj Narang,
  Linxi Fan, Yuke Zhu, and Dieter Fox.
\newblock {{MimicGen}}: A data generation system for scalable robot learning
  using human demonstrations.
\newblock In \emph{Conference on Robot Learning}, 2023.

\bibitem[Mildenhall et~al.(2020)Mildenhall, Srinivasan, Tancik, Barron,
  Ramamoorthi, and Ng]{mildenhall2020nerf}
Ben Mildenhall, Pratul~P. Srinivasan, Matthew Tancik, Jonathan~T. Barron, Ravi
  Ramamoorthi, and Ren Ng.
\newblock {{NeRF}}: Representing scenes as neural radiance fields for view
  synthesis.
\newblock In \emph{European Conference on Computer Vision}, pp.\  405--421,
  2020.

\bibitem[Nasiriany et~al.(2024)Nasiriany, Maddukuri, Zhang, Parikh, Lo, Joshi,
  Mandlekar, and Zhu]{nasiriany2024robocasa}
Soroush Nasiriany, Abhiram Maddukuri, Lance Zhang, Adeet Parikh, Aaron Lo,
  Abhishek Joshi, Ajay Mandlekar, and Yuke Zhu.
\newblock {{RoboCasa}}: Large-scale simulation of household tasks for
  generalist robots.
\newblock In \emph{Robotics: Science and Systems}, 2024.
\newblock \doi{10.15607/RSS.2024.XX.050}.

\bibitem[{NVIDIA}(2023)]{isaacsim2023}
{NVIDIA}.
\newblock {NVIDIA Isaac Sim} documentation.
\newblock Online documentation, 2023.
\newblock URL \url{https://docs.isaacsim.omniverse.nvidia.com/}.
\newblock Accessed: 2026-05-07.

\bibitem[{NVIDIA} et~al.(2025){NVIDIA}, Agarwal, Ali, Bala, Balaji, Barker,
  Cai, Chattopadhyay, Chen, Cui, Ding, Dworakowski, Fan, Fenzi, Ferroni,
  Fidler, Fox, Ge, Ge, Gu, Gururani, He, Huang, Huffman, Jannaty, Jin, Kim,
  Kl{\'a}r, Lam, Lan, Leal-Taixe, Li, Li, Lin, Lin, Ling, Liu, Liu, Luo, Ma,
  Mao, Mo, Mousavian, Nah, Niverty, Page, Paschalidou, Patel, Pavao,
  Ramezanali, Reda, Ren, Sabavat, Schmerling, Shi, Stefaniak, Tang, Tchapmi,
  Tredak, Tseng, Varghese, Wang, Wang, Wang, Wang, Wei, Wei, Wu, Xu, Yang,
  Yen-Chen, Zeng, Zeng, Zhang, Zhang, Zhang, Zhao, and
  Zolkowski]{agarwal2025cosmos}
{NVIDIA}, Niket Agarwal, Arslan Ali, Maciej Bala, Yogesh Balaji, Erik Barker,
  Tiffany Cai, Prithvijit Chattopadhyay, Yongxin Chen, Yin Cui, Yifan Ding,
  Daniel Dworakowski, Jiaojiao Fan, Michele Fenzi, Francesco Ferroni, Sanja
  Fidler, Dieter Fox, Songwei Ge, Yunhao Ge, Jinwei Gu, Siddharth Gururani,
  Ethan He, Jiahui Huang, Jacob Huffman, Pooya Jannaty, Jingyi Jin, Seung~Wook
  Kim, Gergely Kl{\'a}r, Grace Lam, Shiyi Lan, Laura Leal-Taixe, Anqi Li,
  Zhaoshuo Li, Chen-Hsuan Lin, Tsung-Yi Lin, Huan Ling, Ming-Yu Liu, Xian Liu,
  Alice Luo, Qianli Ma, Hanzi Mao, Kaichun Mo, Arsalan Mousavian, Seungjun Nah,
  Sriharsha Niverty, David Page, Despoina Paschalidou, Zeeshan Patel, Lindsey
  Pavao, Morteza Ramezanali, Fitsum Reda, Xiaowei Ren, Vasanth Rao~Naik
  Sabavat, Ed~Schmerling, Stella Shi, Bartosz Stefaniak, Shitao Tang, Lyne
  Tchapmi, Przemek Tredak, Wei-Cheng Tseng, Jibin Varghese, Hao Wang, Haoxiang
  Wang, Heng Wang, Ting-Chun Wang, Fangyin Wei, Xinyue Wei, Jay~Zhangjie Wu,
  Jiashu Xu, Wei Yang, Lin Yen-Chen, Xiaohui Zeng, Yu~Zeng, Jing Zhang,
  Qinsheng Zhang, Yuxuan Zhang, Qingqing Zhao, and Artur Zolkowski.
\newblock {{Cosmos}} world foundation model platform for physical {AI}, 2025.
\newblock URL \url{https://arxiv.org/abs/2501.03575}.

\bibitem[O'Neill et~al.(2024)O'Neill, Rehman, Gupta, Maddukuri, Gupta,
  Padalkar, Lee, Pooley, Gupta, Mandlekar, Jain, Tung, Bewley, Herzog, Irpan,
  Khazatsky, Rai, Gupta, Wang, et~al.]{oneill2024openx}
Abby O'Neill, Abdul Rehman, Abhinav Gupta, Abhiram Maddukuri, Ajay Gupta,
  Abhishek Padalkar, Abraham Lee, Alex Pooley, Anirudh Gupta, Ajay Mandlekar,
  Ashwin Jain, Austin Tung, Alex Bewley, Alexander Herzog, Alex Irpan,
  Alexander Khazatsky, Aviral Rai, Avinash Gupta, Brian Wang, et~al.
\newblock {{Open X-Embodiment}}: Robotic learning datasets and {RT-X} models.
\newblock In \emph{2024 IEEE International Conference on Robotics and
  Automation (ICRA)}, 2024.

\bibitem[Patel et~al.(2025)Patel, Mohan, Mai, Jain, Lazebnik, and
  Li]{patel2025rigvid}
Shivansh Patel, Shraddhaa Mohan, Hanlin Mai, Unnat Jain, Svetlana Lazebnik, and
  Yunzhu Li.
\newblock Robotic manipulation by imitating generated videos without physical
  demonstrations, 2025.
\newblock URL \url{https://arxiv.org/abs/2507.00990}.

\bibitem[Pathak et~al.(2018)Pathak, Mahmoudieh, Luo, Agrawal, Chen, Shentu,
  Shelhamer, Malik, Efros, and Darrell]{pathak2018zeroshot}
Deepak Pathak, Parsa Mahmoudieh, Guanghao Luo, Pulkit Agrawal, Dian Chen, Fred
  Shentu, Evan Shelhamer, Jitendra Malik, Alexei~A. Efros, and Trevor Darrell.
\newblock Zero-shot visual imitation.
\newblock In \emph{International Conference on Learning Representations}, 2018.

\bibitem[Peng et~al.(2018)Peng, Andrychowicz, Zaremba, and Abbeel]{peng2018sim}
Xue~Bin Peng, Marcin Andrychowicz, Wojciech Zaremba, and Pieter Abbeel.
\newblock {Sim-to-Real} transfer of robotic control with dynamics
  randomization.
\newblock In \emph{IEEE International Conference on Robotics and Automation},
  pp.\  3803--3810, 2018.

\bibitem[{Physical Intelligence} et~al.(2025){Physical Intelligence}, Black,
  Brown, Darpinian, Dhabalia, Driess, Esmail, Equi, Finn, Fusai, Galliker,
  Ghosh, Groom, Hausman, Ichter, Jakubczak, Jones, Ke, LeBlanc, Levine,
  Li-Bell, Mothukuri, Nair, Pertsch, Ren, Shi, Smith, Springenberg, Stachowicz,
  Tanner, Vuong, Walke, Walling, Wang, Yu, and Zhilinsky]{pi2025pi05}
{Physical Intelligence}, Kevin Black, Noah Brown, James Darpinian, Karan
  Dhabalia, Danny Driess, Adnan Esmail, Michael Equi, Chelsea Finn, Niccolo
  Fusai, Manuel~Y. Galliker, Dibya Ghosh, Lachy Groom, Karol Hausman, Brian
  Ichter, Szymon Jakubczak, Tim Jones, Liyiming Ke, Devin LeBlanc, Sergey
  Levine, Adrian Li-Bell, Mohith Mothukuri, Suraj Nair, Karl Pertsch, Allen~Z.
  Ren, Lucy~Xiaoyang Shi, Laura Smith, Jost~Tobias Springenberg, Kyle
  Stachowicz, James Tanner, Quan Vuong, Homer Walke, Anna Walling, Haohuan
  Wang, Lili Yu, and Ury Zhilinsky.
\newblock {$\pi_{0.5}$}: A vision-language-action model with open-world
  generalization, 2025.
\newblock URL \url{https://arxiv.org/abs/2504.16054}.

\bibitem[{SAM 3D Team} et~al.(2025){SAM 3D Team}, Chen, Chu, Gleize, Liang,
  Sax, Tang, Wang, Guo, Hardin, Li, Lin, Liu, Ma, Sagar, Song, Wang, Yang,
  Zhang, Doll{\'a}r, Gkioxari, Feiszli, and Malik]{chen2025sam}
{SAM 3D Team}, Xingyu Chen, Fu-Jen Chu, Pierre Gleize, Kevin~J. Liang,
  Alexander Sax, Hao Tang, Weiyao Wang, Michelle Guo, Thibaut Hardin, Xiang Li,
  Aohan Lin, Jiawei Liu, Ziqi Ma, Anushka Sagar, Bowen Song, Xiaodong Wang,
  Jianing Yang, Bowen Zhang, Piotr Doll{\'a}r, Georgia Gkioxari, Matt Feiszli,
  and Jitendra Malik.
\newblock {{SAM 3D}}: {3Dfy} anything in images, 2025.
\newblock URL \url{https://arxiv.org/abs/2511.16624}.

\bibitem[Sch{\"o}nberger \& Frahm(2016)Sch{\"o}nberger and
  Frahm]{schoenberger2016sfm}
Johannes~L. Sch{\"o}nberger and Jan-Michael Frahm.
\newblock Structure-from-motion revisited.
\newblock In \emph{IEEE Conference on Computer Vision and Pattern Recognition},
  pp.\  4104--4113, 2016.

\bibitem[Tobin et~al.(2017)Tobin, Fong, Ray, Schneider, Zaremba, and
  Abbeel]{tobin2017domain}
Josh Tobin, Rachel Fong, Alex Ray, Jonas Schneider, Wojciech Zaremba, and
  Pieter Abbeel.
\newblock Domain randomization for transferring deep neural networks from
  simulation to the real world.
\newblock In \emph{IEEE/RSJ International Conference on Intelligent Robots and
  Systems}, pp.\  23--30, 2017.

\bibitem[Wang et~al.(2024{\natexlab{a}})Wang, Ling, Yuan, Shridhar, Bao, Qin,
  Wang, Xu, and Wang]{wang2023gensim}
Lirui Wang, Yiyang Ling, Zhecheng Yuan, Mohit Shridhar, Chen Bao, Yuzhe Qin,
  Bailin Wang, Huazhe Xu, and Xiaolong Wang.
\newblock {{GenSim}}: Generating robotic simulation tasks via large language
  models.
\newblock In \emph{International Conference on Learning Representations},
  2024{\natexlab{a}}.

\bibitem[Wang et~al.(2024{\natexlab{b}})Wang, Xian, Chen, Wang, Wang,
  Fragkiadaki, Erickson, Held, and Gan]{wang2023robogen}
Yufei Wang, Zhou Xian, Feng Chen, Tsun-Hsuan Wang, Yian Wang, Katerina
  Fragkiadaki, Zackory Erickson, David Held, and Chuang Gan.
\newblock {{RoboGen}}: Towards unleashing infinite data for automated robot
  learning via generative simulation.
\newblock In \emph{Proceedings of the 41st International Conference on Machine
  Learning}, volume 235 of \emph{Proceedings of Machine Learning Research},
  pp.\  51936--51983. PMLR, 2024{\natexlab{b}}.

\bibitem[Wen et~al.(2024)Wen, Yang, Kautz, and
  Birchfield]{wen2024foundationpose}
Bowen Wen, Wei Yang, Jan Kautz, and Stan Birchfield.
\newblock {{FoundationPose}}: Unified {6D} pose estimation and tracking of
  novel objects.
\newblock In \emph{IEEE/CVF Conference on Computer Vision and Pattern
  Recognition}, 2024.

\bibitem[Wu et~al.(2025)Wu, Zhang, Liang, Han, and Dong]{wu2025foundation}
Tianshu Wu, Jiyao Zhang, Shiqian Liang, Zhengxiao Han, and Hao Dong.
\newblock Foundation feature-driven online end-effector pose estimation: A
  marker-free and learning-free approach, 2025.
\newblock URL \url{https://arxiv.org/abs/2503.14051}.

\bibitem[Xiang et~al.(2020)Xiang, Qin, Mo, Xia, Zhu, Liu, Liu, Jiang, Yuan,
  Wang, Yi, Chang, Guibas, and Su]{xiang2020sapien}
Fanbo Xiang, Yuzhe Qin, Kaichun Mo, Yikuan Xia, Hao Zhu, Fangchen Liu, Minghua
  Liu, Hanxiao Jiang, Yifu Yuan, He~Wang, Li~Yi, Angel~X. Chang, Leonidas~J.
  Guibas, and Hao Su.
\newblock {{SAPIEN}}: A {SimulAted} part-based interactive {ENvironment}.
\newblock In \emph{IEEE/CVF Conference on Computer Vision and Pattern
  Recognition}, pp.\  11097--11107, 2020.

\bibitem[Xu et~al.(2025)Xu, Zhang, Hou, Xu, Fan, Veloso, and
  Song]{xu2025dexumi}
Mengda Xu, Han Zhang, Yifan Hou, Zhenjia Xu, Linxi Fan, Manuela Veloso, and
  Shuran Song.
\newblock Dexumi: Using human hand as the universal manipulation interface for
  dexterous manipulation.
\newblock \emph{arXiv preprint arXiv:2505.21864}, 2025.

\bibitem[Xue et~al.(2025)Xue, Deng, Chen, Wang, Yuan, and Xu]{xue2025demogen}
Zhengrong Xue, Shuying Deng, Zhenyang Chen, Yixuan Wang, Zhecheng Yuan, and
  Huazhe Xu.
\newblock {{DemoGen}}: Synthetic demonstration generation for data-efficient
  visuomotor policy learning.
\newblock In \emph{Robotics: Science and Systems}, 2025.

\bibitem[Yang et~al.(2024)Yang, Du, Ghasemipour, Tompson, Schuurmans, and
  Abbeel]{yang2023unisim}
Sherry Yang, Yilun Du, Seyed Kamyar~Seyed Ghasemipour, Jonathan Tompson, Dale
  Schuurmans, and Pieter Abbeel.
\newblock Learning interactive real-world simulators.
\newblock In \emph{International Conference on Learning Representations}, 2024.

\bibitem[Zhu et~al.(2020)Zhu, Wong, Mandlekar, Mart{\'i}n-Mart{\'i}n, Joshi,
  Nasiriany, and Zhu]{zhu2020robosuite}
Yuke Zhu, Josiah Wong, Ajay Mandlekar, Roberto Mart{\'i}n-Mart{\'i}n, Abhishek
  Joshi, Soroush Nasiriany, and Yifeng Zhu.
\newblock {{robosuite}}: A modular simulation framework and benchmark for robot
  learning.
\newblock \emph{arXiv preprint arXiv:2009.12293}, 2020.

\bibitem[Zitkovich et~al.(2023)Zitkovich, Yu, Xu, Xu, Xiao, Xia, Wu, Wohlhart,
  Welker, Wahid, Vuong, Vanhoucke, Tran, Soricut, Singh, Singh, Sermanet,
  Sanketi, Ryoo, Reymann, Rao, Pertsch, Mordatch, Michalewski, Lu, Levine, Lee,
  Leal, Kuang, Julian, Joshi, Irpan, Ichter, Hsu, Herzog, Hausman,
  Gopalakrishnan, Florence, Finn, Dubey, Driess, Ding, Choromanski, Chen,
  Chebotar, Carbajal, Brown, Brohan, Arenas, and Han]{zitkovich2023rt2}
Brianna Zitkovich, Tianhe Yu, Sichun Xu, Peng Xu, Ted Xiao, Fei Xia, Jialin Wu,
  Paul Wohlhart, Stefan Welker, Ayzaan Wahid, Quan Vuong, Vincent Vanhoucke,
  Huong Tran, Radu Soricut, Anikait Singh, Jaspiar Singh, Pierre Sermanet,
  Pannag Sanketi, Michael Ryoo, Krista Reymann, Kanishka Rao, Karl Pertsch,
  Igor Mordatch, Henryk Michalewski, Yao Lu, Sergey Levine, Kuang-Huei Lee,
  Isabel Leal, Yuheng Kuang, Ryan Julian, Nikhil Joshi, Alex Irpan, Brian
  Ichter, Jasmine Hsu, Alexander Herzog, Karol Hausman, Keerthana
  Gopalakrishnan, Pete Florence, Chelsea Finn, Avinava Dubey, Danny Driess,
  Tianli Ding, Krzysztof Choromanski, Xi~Chen, Yevgen Chebotar, Justice
  Carbajal, Noah Brown, Anthony Brohan, Montserrat~Gonzalez Arenas, and Kehang
  Han.
\newblock {{RT-2}}: {Vision-Language-Action} models transfer web knowledge to
  robotic control.
\newblock In \emph{Proceedings of the 7th Conference on Robot Learning}, volume
  229 of \emph{Proceedings of Machine Learning Research}, pp.\  2165--2183.
  PMLR, 2023.

\end{thebibliography}
